\RequirePackage{fix-cm}
\documentclass[smallextended]{svjour3}       
\smartqed  
\usepackage[numbers]{natbib}
\usepackage{graphicx}
\usepackage{times}
\usepackage{epsfig}
\usepackage[ruled]{algorithm2e}
\usepackage{amsmath}
\usepackage{amssymb}
\usepackage{bm}
\usepackage{color}
\usepackage{multirow}                
\usepackage{subfigure}
\usepackage{authblk}

\def\aa{{\boldsymbol{a}}}
\def\bb{{\boldsymbol{b}}}

\def\ee{{\boldsymbol{e}}}

\def\xx{{\boldsymbol{x}}}

\def\yy{{\boldsymbol{y}}}

\def\subjto{{\mbox{subj. to}}}

\renewcommand{\Re}{\mathbb{R}}

  \makeatletter
    \def\multilimits@{\bgroup
  \Let@
  \restore@math@cr
  \default@tag
 \baselineskip\fontdimen10 \scriptfont\tw@
 \advance\baselineskip\fontdimen12 \scriptfont\tw@
 \lineskip\thr@@\fontdimen8 \scriptfont\thr@@
 \lineskiplimit\lineskip
 \vbox\bgroup\ialign\bgroup\hfil$\m@th\scriptstyle{##}$\hfil\crcr}
    \def\Sb{_\multilimits@}
    \def\endSb{\crcr\egroup\egroup\egroup}
\makeatother

{\begin{list}{}%
    {\setlength{\rightmargin}{0pc}%
    \setlength{\leftmargin}{1pc} }} %
{\end{list}}

\newtheorem{Theorem}{Theorem} 

\begin{document}

\title{Sparse Illumination Learning and Transfer for Single-Sample Face Recognition with Image Corruption and Misalignment \thanks{
The work was supported in part by ARO 63092-MA-II, DARPA
FA8650-11-1-7153, ONR N00014-09-1-0230, NSF CCF09-64215,
NSFC No. 61103134 and 61371192, and the Science Foundation for
Outstanding Young Talent of Anhui Province (BJ2101020001). Corresponding author: Allen Y. Yang. 
}
\thanks{A preliminary version of the results was published in \cite{Zhuang2013-CVPR}.
}
}

\titlerunning{Sparse Illumination Learning and Transfer for Single-Sample Face Recognition}        

\author{Liansheng~Zhuang \and Tsung-Han Chan \and Allen~Y.~Yang \and S.~Shankar~Sastry \and Yi~Ma
}


\institute{L.~Zhuang\at
              University of Science and Technology of China, Heifei, China\\
              \email{lszhuang@ustc.edu.cn}           
           \and
           T.-H,~Chan\at
           Advanced Digital Sciences Center, Singapore\\
           \email{thchan@ieee.org}
           \and
           A.~Yang and S.~Sastry\at
           Department of EECS, University of California, Berkeley, CA 94720\\
           \email{\{yang, sastry\}@eecs.berkeley.edu}
           \and
           Y.~Ma\at
           ShanghaiTech University, Shanghai, China\\
           \email{mayi@shanghaitech.edu.cn}
}

\date{Received: date / Accepted: date}

\maketitle

\begin{abstract}
Single-sample face recognition is one of the most challenging problems in face recognition. We propose a novel algorithm to address this problem based on a sparse representation based classification (SRC) framework. The new algorithm is robust to image misalignment and pixel corruption, and is able to reduce required gallery images to one sample per class. To compensate for the missing illumination information traditionally provided by multiple gallery images, a sparse illumination learning and transfer (SILT) technique is introduced. The illumination in SILT is learned by fitting illumination examples of auxiliary face images from one or more additional subjects with a sparsely-used illumination dictionary. By enforcing a sparse representation of the query image in the illumination dictionary, the SILT can effectively recover and transfer the illumination and pose information from the alignment stage to the recognition stage. Our extensive experiments have demonstrated that the new algorithms significantly outperform the state of the art in the single-sample regime and with less restrictions. In particular, the single-sample face alignment accuracy is comparable to that of the well-known Deformable SRC algorithm using multiple gallery images per class. Furthermore, the face recognition accuracy exceeds those of the SRC and Extended SRC algorithms using hand labeled alignment initialization.

\end{abstract}

\section{Introduction}
\label{intro}
Face recognition is one of the classical problems in computer vision. Given a natural image that may contain a human face, it has been known that the appearance of the face image can be easily affected by many image nuisances, including background illumination, pose, and facial corruption/disguise such as makeup, beard, and glasses. Therefore, to develop a \emph{robust} face recognition system whose performance can be comparable to or even exceed that of human vision, the computer system needs to address at least the following three closely related problems: First, it needs to effectively model the change of illumination on the human face. Second, it needs to align the pose of the face. Third, it needs to tolerance the corruption of facial features that leads to potential gross pixel error against the gallery images.

In the literature, many well-known solutions have been studied to tackle these problems \cite{HagerG1998-PAMI,ZhaoW2003,HoJ2003-CVPR,GaneshA2011}, although a complete review of the field is outside the scope of this paper. More recently, a new face recognition framework called \emph{sparse-representation based classification} (SRC) was proposed \cite{WrightJ2009-PAMI}, which can successfully address most of the above problems. The framework is built on a subspace illumination model characterizing the distribution of a corruption-free face image sample (stacked in vector form) under a fixed pose, one subspace model per subject class \cite{BelhumeurP1997-PAMI,BasriR2003-PAMI}. When an unknown query image is jointly represented by all the subspace models, only a small subset of these subspace coefficients need to be nonzero, which would primarily correspond to the subspace model of the true subject. Therefore, by optimizing the sparsity of such an overcomplete linear representation, the dominant nonzero coefficients indicate the identity of the query image. In the case of image corruption, since the corruption typically only affects a sparse set of pixel values, one can concurrently optimize a sparse error term in the image space to compensate for the corrupted pixel values.

In practice, a face image may appear at any image location with random background. Therefore, a face detection and registration step is typically first used to detect the face image. Most of the methods in face detection would learn a class of local image features/patches that are sensitive to the appearance of key facial features \cite{YanS2003,ViolaP2004-IJCV,LiangL2008-ECCV}. Using either an active shape model \cite{CootesT1995-CVIU} or an active appearance model \cite{CootesT1998-ECCV}, the location of the face can be detected even when the expression of the face is not neutral or some facial features are occluded \cite{SaragihJ2009-ICCV,GuL2008-ECCV}. However, using these face registration algorithms \emph{alone} is not sufficient to align a query image to gallery images in SRC. The main reasons are two-fold: First,  except for some fast detectors such as Viola-Jones \cite{ViolaP2004-IJCV}, more sophisticated detectors are expensive to run and require learning prior distribution of the shape model from meticulously hand-labeled gallery images. More importantly, these detectors would register the pixel values of the query image with respect to the \emph{average} shape model learned from all the gallery images, but they typically cannot align the pixel values of the query image to the gallery images for the purpose of recognition, as required in SRC.

Following the sparse representation framework in \cite{WrightJ2009-PAMI,WagnerA2012-PAMI}, we propose a novel algorithm to effectively extend SRC for face alignment and recognition in the small-sample-set scenario. We observe that in addition to the aforementioned image nuisances, one of the outstanding challenges in face recognition is indeed the small sample set problem. For instance, in many biometric, surveillance, and Internet applications, there may be only a few gallery examples that are collected for a subject of interest, and the subject may not be able to undergo a comprehensive image collection session in a laboratory.\footnote{In this paper, we use Viola-Jones face detector to initialize the face image location. As a result, we do not consider scenarios where the face may contain a large 3D transformation or large expression change. These more severe conditions can be addressed in the face detection stage using more sophisticated face models as we previously mentioned.}

Unfortunately, most of the existing SRC-based alignment and recognition algorithms would fail in such scenarios. For starters, the original SRC algorithm \cite{WrightJ2009-PAMI} assumes a plurality of gallery samples from each class must sufficiently span its illumination subspace. The algorithm performs poorly in the single sample regime, as we will later shown in our experiment. In \cite{WagnerA2012-PAMI}, in order to guarantee that the gallery images contain sufficient illumination patterns, the test subjects must further go through a nontrivial passport-style image collection process in a dark room in order to be entered into the gallery database. More recently, another development in the SRC framework is simultaneous face alignment and recognition methods \cite{YanS2010-TIP,HuangJ2008-CVPR,YangM2012-ECCV}. Nevertheless, these methods did not go beyond the basic assumption used in SRC and other prior art that the face illumination model is measured by multiple gallery samples for each class. Furthermore, as shown in \cite{WagnerA2012-PAMI}, robust face alignment and recognition can be solved separately as a two-step process, as long as the recovered image transformation can be carried over from the alignment stage to the recognition stage. Therefore, simultaneous face alignment and recognition could make the already expensive sparse optimization problem even more difficult to solve.

\subsection{Contributions}
Single-sample face alignment and recognition represents an important step towards practical face recognition solutions using images collected in the wild or on the Internet. We contend that the problem can be solved quite effectively by an elegant algorithm. The key observation is that one sample per class mainly deprives the algorithm of an illumination subspace model for individual classes. We show that an \emph{illumination dictionary} can be learned from additional subject classes to compensate for the lack of the illumination information in the gallery set.

Due to the fact that the variations of human faces are usually smaller than illumination changes of the same face, we propose a dictionary learning method to decompose the face images as vectors into two components: a low-rank matrix encodes the subject identities while a sparsely-used matrix (or dictionary) represents the possible illumination variations. The auxiliary illumination images can be selected outside the set of gallery subjects. Since most of the information associated with the subject identities is contained in the rank-constrained matrix, the sparsely-used illumination dictionary is expected to be subject-invariant. Finally, we show that the other image nuisances, including pose variation and image corruption, can be readily corrected by a single gallery image of \emph{arbitrary illumination condition} combined with the illumination dictionary. The algorithm also does not need to know the information of any possible facial corruption for the algorithm to be robust. The new method is called \emph{sparse illumination learning and transfer} (SILT). Similarly, the illumination dictionary defined in the method will be referred to as the SILT dictionary.

Preliminary results of this work were first reported in our conference paper \cite{Zhuang2013-CVPR}. To the best of our knowledge, the paper \cite{Zhuang2013-CVPR} was the first to propose a solution to perform small-sample-set facial alignment and recognition via a sparse illumination transfer. However, the construction of the illumination dictionary in \cite{Zhuang2013-CVPR} was largely ad hoc via a simple concatenation of the auxiliary illumination samples. It was suggested in \cite{Zhuang2013-CVPR} that a sparse illumination representation can be found to compensate for the missing illumination model in single gallery images. In this paper, we propose a new illumination dictionary model to specifically learn the dictionary from the auxiliary images. We also study efficient optimization algorithms to solve the dictionary learning problem numerically. Finally, more comprehensive experiments are conducted, especially on the case when the number of available illumination learning subjects grows from one to many. In the largest scale, we employ all the 38 available subjects in the Extended YaleB database \cite{LeeK2005-PAMI} as the auxiliary illumination samples. The new results show improved recognition results than those in \cite{Zhuang2013-CVPR}.

In terms of the algorithm complexity, learning the SILT dictionary contains two successive procedures; one is principal component analysis (PCA)-like solution while the other involves solving a sequence of linear programs. The leaning algorithm is almost parameter-free, only dependent on the dictionary size. Applying the SILT dictionary in the alignment and recognition stages potentially can significantly improve the speed of SRC-type algorithms, because a sparse optimization solver such as those in \cite{YangA2012-TIP} is now faced with much smaller linear systems that only involves a single sample per class plus a small learned illumination dictionary.

This paper bears resemblance to the work called Extended SRC \cite{DengW2012-PAMI}, whereby an intraclass variant dictionary was similarly added to be a part of the SRC objective function for recognition. Our work differs from \cite{DengW2012-PAMI} in that the proposed SILT dictionary is automatically learned from a selection of independent subject(s), whereas in \cite{DengW2012-PAMI}, the dictionary is simply hand-crafted. Yet, the subject classes used to learn the SILT dictionary is also impartial to the gallery classes. Furthermore, by transferring both the pose and illumination from the alignment stage to the recognition stage, our algorithm can handle insufficient illumination and misalignment at the same time, and allows for the single reference images to have arbitrary illumination conditions. Finally, our algorithm is also robust to moderate amounts of image pixel corruption, even though we do not need to include any image corruption examples in the SILT dictionary, while in \cite{DengW2012-PAMI} the intraclass variant dictionary uses both normal and corrupted face samples. We also compare our performance with \cite{DengW2012-PAMI} in Section \ref{sec:experiment}.

More recently, the problem of single-sample face recognition was considered in another work \cite{YangM2013-ICCV}, called \emph{sparse variation dictionary learning} (SVDL). The work proposed an alternative method to learn a sparse variation dictionary that amends the SRC framework with single samples. The main difference between the two dictionary learning algorithms is that in SVDL, both the illumination learning images and the gallery images are involved in the dictionary learning algorithm. The authors argued that jointly considering the illumination samples and the gallery samples helps to generate a very compact, adaptive dictionary that exploits the correlation between the illumination learning set and the gallery set. While in this paper, the learning of the SILT dictionary is independent of the gallery set and the alignment and recognition tasks. Therefore, the learned dictionary can be estimated off-line and without costing any computational penalty when the gallery images are presented. Furthermore, the SILT framework addresses both face alignment and recognition problems, and is capable of transferring both the illumination \emph{and} pose information from the alignment stage to the recognition stage. In contrast, SVDL in \cite{YangM2013-ICCV} only concerns face recognition with a frontal position, and its complexity would grow substantially when its adaptive dictionary needs to be re-computed under varying poses of the query image. We will show in Section \ref{sec:experiment} that, without considering this pose-related computational penalty for SVDL, the SILT framework outperforms SVDL in both recognition accuracy and robustness to pixel corruption.

\section{Sparse Representation-based Classification}
\label{sec:SRC}
In this section, we first briefly review the SRC framework. Assume a face image $\bb\in\Re^d$ in grayscale can be written in vector form by stacking its pixels. Given $L$ subject classes, assume $n_i$ well-aligned gallery images $A_i = [\aa_{i,1}, \aa_{i,2}, \cdots, \aa_{i, n_i}]\in\Re^{d\times n_i}$ of the same dimension as $\bb$ are sampled for the $i$-th class under the frontal position and various illumination conditions. These gallery images are further aligned in terms of the coordinates of some salient facial features, e.g., eye corners and/or mouth corners. For brevity, the gallery images under such conditions are said to be in the \emph{neutral position}. Furthermore, we do not explicitly model the variation of facial expression in this paper. Based on the illumination subspace assumption, if $\bb$ belongs to the $i$-th class, then $\bb$ lies in the low-dimensional subspace spanned by the gallery images in $A_i$, namely,
\begin{equation}
\bb = A_i\xx_i.
\end{equation}

When the query image $\bb$ is captured in practice, it may contain an unknown 3D pose that is different from the neutral position. In image registration literature \cite{LucasB1981,HagerG1998-PAMI,WagnerA2012-PAMI}, the effect of the 3D pose can be modeled as an image transformation as $\tau\in T$, where $T$ is a finite-dimensional group of transformations, such as translation, similarity transform, affine transform, and homography. The goal of face alignment is to recover the transformation $\tau$, such that the unwarped query image $\bb_0$ in the neutral position remains in the same illumination subspace: $\bb_0 \doteq \bb\circ \tau = A_i\xx_i$.

In robust face alignment, the issue is often further exacerbated by the cascade of complex illumination patterns and moderate image pixel corruption and occlusion. In the SRC framework \cite{WrightJ2009-PAMI,WagnerA2012-PAMI}, the combined effect of image misalignment and sparse corruption is modeled by
\begin{equation}
\hat{\tau}_i = \arg\min_{\xx_i, \ee, \tau_i}\|\ee\|_1\quad \subjto \quad \bb\circ \tau_i = A_i\xx_i + \ee,
\label{eq:SRC-alignment}
\end{equation}
where the alignment is achieved on a per-class basis for each $A_i$, and $\ee\in\Re^d$ is the sparse alignment error. After linearizing the potentially nonlinear image transformation function $\tau$, \eqref{eq:SRC-alignment} can be solved iteratively by a standard $\ell_1$-minimization solver. In \cite{WagnerA2012-PAMI}, it was shown that the alignment based on \eqref{eq:SRC-alignment} can tolerate translation shift up to 20\% of the between-eye distance and up to $30^\circ$ in-plane rotation, which is typically sufficient to compensate moderate misalignment caused by a good face detector.

Once the optimal transformation $\tau_i$ is recovered for each class $i$, the transformation is carried over to the recognition algorithm, where the gallery images in each $A_i$ are transformed by $\tau_i^{-1}$ to align with the query image $\bb$. Finally, a global sparse representation $\xx$ with respect to the transformed gallery images is sought by solving the following sparse optimization problem:
\begin{equation}
\begin{array}{rcl}
\xx^* &=& \arg\min_{\xx, \ee} \|\xx\|_1 + \|\ee\|_1 \\
\subjto & & \bb = \left[A_1\circ \tau_1^{-1},\cdots, A_L\circ \tau_L^{-1}\right]\xx + \ee.
\end{array}
\label{eq:CAB}
\end{equation}
One can further show that when the correlation of the face samples in $A$ is sufficiently tight in the high-dimensional image space, solving \eqref{eq:CAB} via $\ell_1$-minimization
guarantees to recover both the sparse coefficients $\xx$ and very dense (sparsity $\rho \nearrow 1$) randomly signed error $\ee$ \cite{WrightJ2010-IT}.

\section{Sparse Illumination Learning and Transfer}
\label{sec:SILT}
In this section, we propose a novel face alignment algorithm that is effective even when a very small number of training images are provided per class, called \emph{sparse illumination learning and transfer} (SILT). In the extreme case, we specifically consider the \emph{single-sample face alignment problem} where only one training image $\aa_i$ of \emph{arbitrary illumination} is available from class $i$. The same algorithm easily extends to the case when multiple training images are provided. In Section \ref{sec:recognition}, we will show how to integrate the estimation of SILT in robust single-sample face recognition. In Section \ref{sec:experiment}, we further show in our experiment that SILT is also complementary and useful in other existing face recognition methods as an image pre-processing step.

\subsection{Illumination Dictionary Learning}
To mitigate the scarcity of the training images, something has to give to recover the missing illumination model under which the image appearance of a human face can be affected. Motivated by the idea of transfer learning \cite{DoC2005-NIPS,QuattoniA2008-CVPR,LampertC2009-CVPR}, we stipulate that one can obtain the illumination information for both alignment \emph{and} recognition from a set of additional subject classes, called the \emph{illumination dictionary}. The auxiliary face images for learning the illumination dictionary have the same frontal pose as the gallery images, and can be collected offline and different from the query classes $A=[A_1, \cdots, A_L]$. In other words, no matter how scarce the gallery images are, one can always obtain a potentially large set of auxiliary face images from other unrelated subjects who may have similar face shapes as the query subjects and may provide sufficient illumination examples.

Suppose that we are given face images of sufficient illumination patterns for additional $P$ subjects $D = [D_1,\cdots, D_p] \in \Re^{d\times(np)}$, and assume without loss of generality that each subject contains $n$ face images, i.e., $D_i\in \Re^{d\times n}$ for subject $i$, and each image has the same dimension as the gallery images. 

Our hope is that $D$ can be expressed by a superposition of a rank-constrained matrix and a sparsely-used matrix:
\begin{equation}\label{eq: L+BS}
D = V \otimes {\mathbf 1}^T + C S,
\end{equation}
where $V \in \Re^{d\times p}$ is a matrix where each column vector represents a subject class from 1 to $p$, $\mathbf 1\in\Re^n$, $C\in \Re^{d\times k}$ is a learned illumination dictionary, and $S\in \Re^{k \times np}$ is a sparse matrix. Here, $\otimes$ denotes the Kronecker product, and hence the first term $V \otimes {\mathbf 1}^T\in\Re^{d\times(np)}$ in \eqref{eq: L+BS} is clearly low rank. We also assume that $k\leq \min\{d,np\}$ for $C$ to prevent model over-fitting. 

One can better understand the roles of the different matrices in \eqref{eq: L+BS} as follows: $V \otimes {\mathbf 1}^T$ describes the inter-class variation associated with the $p$ different subject identifies, $C$ describes the common intra-class variation associated with the illumination change, and $S$ operates like a sparse representation of illumination patterns that compensate the singular subject images in $V$. Considering  other possible face variations, we may further add a small error term $E \in \Re^{d\times np}$ in \eqref{eq: L+BS} as
\begin{equation}\label{eq: L+BS+E}
D = V \otimes {\mathbf 1}^T + C S + E.
\end{equation}

To encourage sparsity of $S$ and minimum fitting error $E$, we formulate the illumination dictionary learning problem as an optimization problem
\begin{equation}\label{eq: modeling_ori}
\begin{split} \min_{V,C,S,E}  \|S\|_{0} + \|E\|_{F}   ~~ \subjto~ D = V \otimes {\mathbf 1}^T + CS + E,
\end{split}
\end{equation}
where $\|\cdot\|_0$ denotes the matrix $\ell_0$-norm and $\|\cdot\|_{F}$ is the Frobenius norm. Note that in the SRC framework such as \eqref{eq:SRC-alignment} and \eqref{eq:CAB}, the image corruption has been traditionally estimated by minimizing a sparse error term $\| E \|_0$. The reason we can model a dense error term using $\|E\|_{F}$ is that the selection of the auxiliary illumination examples is conducted manually and offline. Therefore, it is reasonable to assume that the face images in $D$ do not contain significant facial disguise and pixel corruption. This assumption also simplifies the complexity of the optimization problem in \eqref{eq: modeling_ori}.

\subsection{Numerical Implementation}
Solving \eqref{eq: modeling_ori} is a challenging problem, mainly because it has a non-convex objective function \emph{and} a non-convex, non-linear constraint. In optimization, the standard procedure to relax the non-convex objective function is to find a good convex surrogate. However, the second problem about how to handle the non-convex, non-linear constraint is less understood. Although the well-known alternating direction method \cite{GabayD1976,TsengP1991,BoydS2011} can be applied, the solution may not converge to the global optimum.

In the following, we will reformulate the constraint in \eqref{eq: modeling_ori} and propose a successive optimization algorithm. The algorithm can be shown numerically to recover $V,C,S,E$ exactly if $S$ is sufficiently sparse.

First, we reformulate the constraint of \eqref{eq: modeling_ori} as follows:
\begin{align}
{D} &= \underbrace{({V}-{C}{F})}_{\overline{V}}\otimes{\mathbf 1}^T + \underbrace{{C}{W}}_{\overline{C}}\underbrace{{W}^{-1}({S}+{F}\otimes{\mathbf 1}^T)}_{H} +{E}, \label{eq: details}
\end{align}
where ${F}\in\mathbb{R}^{k\times p}$ measures the possible ambiguity between the first two terms of the right hand side, and ${W} \in\mathbb{R}^{k\times k}$ is a non-singular transformation such that $\overline{{C}}^T\overline{{C}}={I}$, where ${I}$ is the identity matrix of proper dimension. 

From \eqref{eq: details}, we have
\begin{equation}\label{eq: change_variable}
{S} = {W}{H} - {F}\otimes{\mathbf 1}^T.
\end{equation}
Hence, problem \eqref{eq: modeling_ori} can be written as:
\begin{equation}\label{eq: modeling_reformulated_successive}
\min\begin{Sb} {{\rm rank}({W})=k}\\{{F}}\end{Sb} \Big[\|{W}{H} - {F}\otimes{\mathbf 1}^T\|_{0} +
\big(\min\begin{Sb}{{D} = \overline{{V}}\otimes{\mathbf 1}^T
+ \overline{{C}}{H} + {E}}\\{\overline{{C}}^T\overline{{C}} = {I}}\end{Sb} ~~~ \|{E}\|_{F} \big)\Big].
\end{equation}
The new formulation in \eqref{eq: modeling_reformulated_successive} allows us to apply a
successive optimization strategy. In this case, successive optimization exploits the successive structure of \eqref{eq: modeling_reformulated_successive} to recursively approximate problem \eqref{eq: modeling_ori}. Although it is a heuristic, but it can have promising performance in practice.

More specifically, we approximate
problem \eqref{eq: modeling_reformulated_successive} by decoupling it into two successively processed stages:
\begin{align}
\{\overline{{V}}^*,\overline{{C}}^*, {H}^*,
&{E}^*\} = {\rm arg}\min ~\|{E}\|_{F}\label{eq: successive1}\\
& ~\subjto~~{D} = \overline{{V}}\otimes{\mathbf
1}^T + \overline{{C}}{H} + {E},~\overline{{C}}^T\overline{{C}} = {I}, \nonumber\\
\{{W}^*,&{F}^*\}  =  {\rm arg}\min ~\|{W}{H}^* - {F}\otimes{\mathbf 1}^T\|_{0}\label{eq: successive2}\\
& ~~~~~~~~~\subjto~~{\rm rank}({W})=k. \nonumber
\end{align}
Suppose that $\{\overline{{V}}^*,\overline{{C}}^*, {H}^*, {E}^*,{W}^*,{F}^*\}$ are found, then the solutions of the other variables in problem \eqref{eq: modeling_ori} are given by
\begin{subequations}\label{eq: post_process}\begin{align}
{C}^* &= \overline{{C}}^*({W}^*)^{-1},\label{eq: postprocessing1} \\
{V}^* &= \overline{{V}}^* + {C}^*{F}^*,\label{eq: postprocessing2}\\
{S}^* & = {W}^*{H}^* - {F}^*\otimes{\mathbf 1}^T. \label{eq: postprocessing3}
\end{align}
\end{subequations}

In what follows, we describe how to solve problem \eqref{eq: successive1} and \eqref{eq: successive2}.
\subsubsection{Solving Problem \eqref{eq: successive1}}
Problem \eqref{eq: successive1} is a difficult non-convex problem. Fortunately we can prove that it has a closed-form solution, as stated in the following theorem:
\begin{Theorem}Suppose that $D = [D_1,D_2,...,D_p]$ where $D_i$ is the training set associated with subject $i$, and assume each subject has $n$ images. Problem \eqref{eq: successive1} has the following closed-form solution:
\begin{eqnarray}\label{eq: PCA}
\begin{array}{rcl}
\overline{{V}}^* &=& [~\frac{1}{n}{D}_1{\mathbf 1},\frac{1}{n}{D}_2{\mathbf
1},\hdots,\frac{1}{n}{D}_p{\mathbf 1}~], \label{eq: Av2}\\
\overline{{C}}^* &=& [~ \boldsymbol{q}_1( {U}{U}^T
), \boldsymbol{q}_2( {U}{U}^T ), \hdots, \boldsymbol{q}_{k}(
{U}{U}^T ) ~], \label{eq: B_bar}\\
{H}^* &=& (\overline{{C}}^*)^T({D} -
\overline{{V}}^*\otimes{\mathbf 1}^T),\\
{E}^* &=& {D} - \overline{{V}}^*\otimes{\mathbf
1}^T - \overline{{C}}^*{H}^*.
\end{array}
\end{eqnarray}
where ${U}={D} - \overline{{V}}^*\otimes{\mathbf 1}^T$ and ${\boldsymbol{q}_i({Z})}$ is the eigenvector associated with the
$i$th principal eigenvalue of the square matrix ${Z}$.
\label{thm}
\end{Theorem}

The proof of Theorem \ref{thm} is given in Appendix. To better understand the closed-form solution, we can see that each column of $\overline{{V}}^*$ represents the mean vector of a training set $D_i$. Therefore, ${U}$ represents a normalized data matrix when the mean vectors are removed from $D$. Since the column vectors of $\overline{{C}}$ are the first $k$ orthonomal basis vectors that maximizes the inter-class variance, it can be thought of as a variant of principal component analysis (PCA).

\subsubsection{Solving Problem \eqref{eq: successive2}}
We now turn our attention to problem \eqref{eq: successive2}, which is more difficult than \eqref{eq: successive1}. The problem is very similar to a conventional sparse dictionary learning problem, where the goal is to learn a basis that most compactly represents the face images $D$. While many heuristics have been proposed before (\emph{e.g.}, see \cite{Aharon2006} and the references therein), because of its combinatorial nature, this problem is difficult to solve efficiently. 

Our solution of \eqref{eq: successive2} is largely inspired by a recent paper \cite{Daniel2012}, which shows that the inverse problem can be well-defined, and there exist efficient and provably correct algorithms to solve the inverse problem. The only difference lies in that our problem has an additional unknown matrix ${F}$ here. Hence, we propose to solve problem \eqref{eq: successive2} by solving the following linear programs sequentially; that is, for $i$ from 1 to $k$, we solve
\begin{equation}\label{eq:
modeling11111}
\begin{split}
 \{\hat{\bm{w}}_i, \hat{\bm{f}}_i\} = {\rm
arg} &\min_{\bm{w}\in\mathbb{R}^k,\bm{f}\in\mathbb{R}^p}~\|\bm{w}^T {H}^* - \bm{f}^T\otimes{\mathbf 1}^T\|_1,\\
&~\subjto~\bm{w}^T{P}^\perp_{\hat{{W}}_{i-1}}{\mathbf r} =
1,
\end{split}
\end{equation}where $\hat{\bm{w}}_i^T$ and $\hat{\bm{f}}_i^T$ denote the estimates of the $i$th
row vector of ${W}$ and ${F}$, respectively, $\hat{{W}}_{i-1} = [\hat{\bm{w}}_1,...,\hat{\bm{w}}_{i-1}]\in\mathbb{R}^{k\times (i-1)}$ denotes a matrix comprising previously found solutions, ${P}_{\hat{{W}}_{i-1}}^\perp$ is the orthogonal complement projector of
$\hat{{W}}_{i-1}$, and ${\mathbf r}\in\mathbb{R}^{k}$ is an analysis filter. Note that the constraint in \eqref{eq:
modeling11111} is to ensure ${P}_{\hat{{W}}_{i-1}}^{\perp} \bm{w} \neq {\mathbf 0},~\forall~i$, and so the rank
of the final solution ${W}^*$ is equal to $k$.

The intuition behind \eqref{eq: modeling11111} is to use a sequence of $\ell_1$-minimization (or linear programs) to approximate the non-convex $\ell_0$ minimization problem \eqref{eq: successive2}. While the problem addressed in \cite{Daniel2012} slightly differs from \eqref{eq: modeling11111}, their theoretical results may suggest us how to choose the analysis filter ${\mathbf r}$. Applying their results to our problem, we select ${\mathbf r}$ to be a column of ${H}^*$ and choose the solution to be the one that results in minimum cardinality.

The details of the successive optimization for problem \eqref{eq: modeling_ori} are summarized in Algorithm \ref{algorithm_SOP}. Here, $[\cdot]_i$ denotes the $i$th column of a matrix. Note that the proposed method only has the number of atoms $k$ to tune. Therefore, it generates consistent results for a given dataset and $k$. 

\begin{algorithm}
\SetKwInOut{Input}{input} \SetKwInOut{Output}{output}
\Input{Data matrix $D$, and number of atoms $k$.}

{\bf initialize} $\hat{{W}} = {\mathbf 0}$ and
$\hat{{F}} = {\mathbf 0}$.

compute $\overline{{V}}^*$, $\overline{{C}}^*$, ${H}^*$, and ${E}^*$ by
\eqref{eq: PCA}.

\For{$i=1,...,k$} {

\For{$j=1,...,n$} {

choose ${\mathbf r} = [{H}^*]_j$.

compute $\{\hat{\bm{w}}_{ij}, \hat{\bm{f}}_{ij}\}$ by \eqref{eq:
modeling11111}.} 

compute $\ell \in {\rm arg} \min_{j} \|\hat{\bm{w}}^T_{ij} {H}^*
- \hat{\bm{f}}^T_{ij}\otimes{\mathbf 1}^T \|_0$.

update $(\hat{\bm{w}}_i,\hat{\bm{f}}_i) = (\hat{\bm{w}}_{i\ell},\hat{\bm{f}}_{i\ell})$.

update $[\hat{{W}}]_i = \hat{\bm{w}}_i$ and
$[\hat{{F}}]_i = \hat{\bm{f}}_i$.}

update $({W}^*,{F}^*) = (\hat{{W}}^T,\hat{{F}}^T)$.

compute ${C}^*$, ${V}^*$, ${S}^*$ by \eqref{eq: post_process}.

\Output{solution $({V}^*, {C}^*, {S}^*, {E}^*)$.} \caption{Successive optimization for \eqref{eq: modeling_ori}.}
\label{algorithm_SOP}
\end{algorithm}

\begin{example}
To illustrate the illumination dictionary model in \eqref{eq: L+BS+E}, we conduct a simple experiment on Extended YaleB database \cite{LeeK2005-PAMI}. Only the frontal images of the 38 subjects in the database are included. Figure \ref{fig:dictoinaries} illustrates the learned identity vectors of the first 10 subjects in $V$ and the first 10 atoms in the illumination dictionary $C$.

\begin{figure}[ht]\centering
\includegraphics[scale = 0.42]{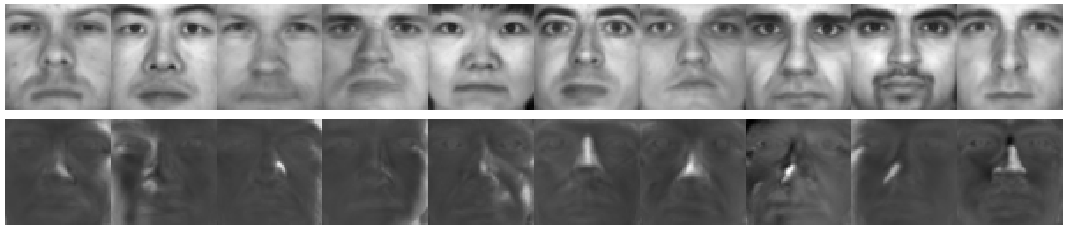}
\caption{{\bf Top:} First ten columns of $V$ unstacked as subject identity images. {\bf Bottom:} First ten columns of $C$ unstacked as illumination images. }
\label{fig:dictoinaries}
\end{figure}
\end{example}

In the next section, we will propose an extension of the SRC framework using SILT, which is aimed at addressing both alignment and recognition with small gallery samples. In particular, among the estimates from Algorithm \ref{algorithm_SOP}, only the illumination dictionary $C$ will be used in the subsequent sparse illumination transfer process. We should emphasize here that in the literature, there are several other algorithms that deal with illumination transfer functions, such as the quotient image \cite{ShashuaA2001-PAMI,PeersP2007} and edge-preserving filters \cite{ChenX2011-CVPR}. The focus of this paper is to learn an illumination dictionary for single-sample alignment and recognition in the SRC framework. The approach of adding an auxiliary dictionary to help recognition was also considered in \cite{DengW2012-PAMI,YangM2013-ICCV}. However, most of these illumination transfer methods are only for recognition but not alignment.

\section{Robust Single-Sample Face Alignment and Recognition using SILT}
\subsection{Robust Single-Sample Alignment}
\label{sec:alignment}

Without loss of generality, we assume each gallery class only contains one sample $A_i = \aa_i$. It is important to note that in our problem setting, each $\aa_i$ can be sampled from an arbitrary lighting condition, and we do not assume the gallery images to share the same illumination pattern. In the alignment stage, given a query image $\bb$, we estimate an image transformation $\tau_i$ applied in the 2-D image coordinates of $\bb$ to align it with $\aa_i$. Clearly, if one were to directly apply the standard SRC solution \eqref{eq:SRC-alignment}, the so-defined alignment error $\ee = \bb\circ \tau_i - \aa_i x_i$ may not be sparse. More specifically, the different illumination conditions between $\bb$ and $\aa_i$ may introduce a \emph{dense} alignment error even when the two images are perfectly aligned. Although an alignment error can still be minimized with respect to an $\ell_1$-norm or $\ell_2$-norm penalty, the algorithm would lose its robustness when concurrently handling sparse image corruption and facial disguise.

\begin{figure}[ht!]
\centering
\includegraphics[width=1.5in]{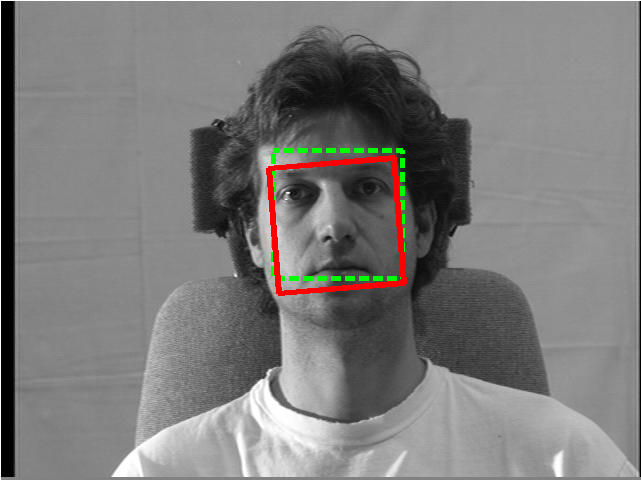}
\includegraphics[width=1.5in]{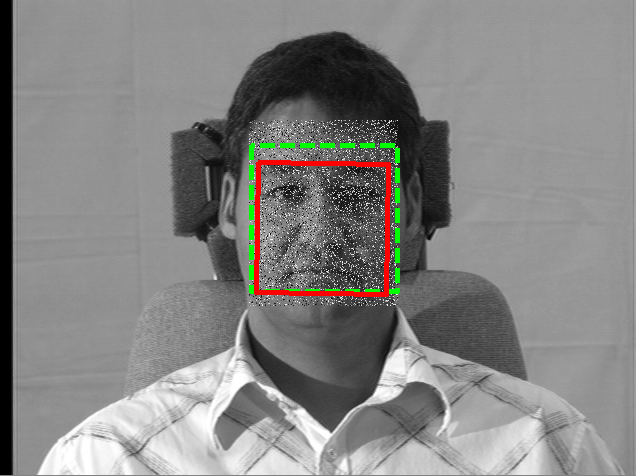}
\caption{Single-sample alignment results on Multi-PIE. The solid red boxes are the initial face locations provided by a face detector. The dash green boxes show the alignment results. The subject image on the right has 30\% of the face pixels corrupted by random noise.}
\label{fig:alignment}
\end{figure}

The SILT algorithm mitigates the problem by using the sparsely-used illumination dictionary $C$ to compensate the illumination difference between $\bb$ and $\aa_i$. More specifically, SILT alignment solves the following problem:
\begin{equation}
\begin{array}{rcl}
(\hat{\tau}_i,\hat{x}_i,\hat{\yy}_i) & = & \arg\min_{\tau_i,x_i,\yy_i,\ee} \|\yy_i\|_1 + \lambda \|\ee\|_1 \\
\subjto & & \bb\circ\tau_i = \aa_ix_i + C\yy_i + \ee.
\end{array}
\label{eq:SILT-alignment}
\end{equation}
In \eqref{eq:SILT-alignment}, $\lambda>0$ is a parameter that balances the weight of $\yy_i$ and $\ee$, which can be chosen empirically. In our experiment, we find $\lambda=1$ generally leads to good performance for both uncorrupted and corrupted cases. $C$ is the SILT dictionary learned in Algorithm \ref{algorithm_SOP}. Finally, the objective function \eqref{eq:SILT-alignment} can be solved efficiently using $\ell_1$-minimization techniques such as those discussed in \cite{WagnerA2012-PAMI,YangA2012-TIP}. Figure \ref{fig:alignment} shows two examples of the SILT alignment results.

\subsection{Robust Single-Sample Recognition}
\label{sec:recognition}

Next, we propose a novel face recognition algorithm that extends the SRC framework to the single-sample case. Similar to the above alignment algorithm, the algorithm also applies trivially when multiple gallery samples are available per class.

In the previous SRC framework \eqref{eq:CAB}, once the transformation $\tau_i$ is recovered for each class $A_i$, the transformation is carried over to the recognition stage, where the gallery images in $A_i$ are transformed by $\tau_i^{-1}$ to align with the query image $\bb$. In the single-sample case, the sparse representation model in \eqref{eq:CAB} will not be satisfied due to two reasons. First, as the $A$ matrix only contains one sample per class, even when $\bb$ is a valid query image with no gross image corruption or facial disguise, the equality constraint $\bb = A\xx$ typically will not hold true. As a result, it becomes difficult to classify $\bb$ based on the sparse coefficients of $\xx$ as suggested in SRC. Second, as the illumination condition of $\bb$ may not be fully expressed by the linear combination $A\xx$, it causes the error $\ee = \bb - A \xx$ to be dense, mostly to compensate the difference in their illumination conditions. The problem reduces the effectiveness of SRC to compensate gross image corruption by minimizing the sparsity of $\ee$.

In the SILT framework, we have seen in \eqref{eq:SILT-alignment} that if an auxiliary illumination dictionary $C$ is provided, it can be used to compensate the missing illumination information in single gallery images $\aa_i$. Therefore, in the recognition stage, one may consider transfer both the illumination information $C\hat{\yy}_i$ and alignment $\tau_i$ to compensate each $\aa_i$:
\begin{equation}
\tilde{\aa}_i = (\aa_i \hat{x}_i + C\hat{\yy}_i)\circ \tau_i^{-1}.
\label{eq:warp-a}
\end{equation}
The collection of all the warped gallery images is defined as $\tilde{A} = [\tilde{\aa}_1, \tilde{\aa}_2, \ldots, \tilde{\aa}_L]$.

Unfortunately, a careful examination of this proposal \eqref{eq:warp-a} reveals a rather subtle issue that prevents us to directly apply the warped gallery images $\tilde{A}$ in the recognition stage. The problem lies in the fact that the illumination dictionary $C$ is learned from the auxiliary face images with the frontal position that are typically cropped and normalized. As a result, the atoms of the dictionary $C$ cannot be simply warped by an image transformation $\tau_i^{-1}$. An exact solution to update the pose of the illumination dictionary $C$ would require the algorithm to first warp the auxiliary images themselves in $D$, and then retrain the illumination dictionary $C_{\tau_i^{-1}}$ for each transformation $\tau_i$. Clearly, this task is prohibitively expensive.\footnote{In our previous work \cite{Zhuang2013-CVPR}, this simple extension was in fact used as the solution to transfer both the alignment and illumination information from the alignment stage to the recognition stage. However, the assumption was valid because the illumination dictionary used in \cite{Zhuang2013-CVPR} was constructed by concatenating the auxiliary images themselves, namely, $D$ in this paper. Therefore, the problem of warping a learned dictionary was mitigated.}

\begin{figure}[th!]
\centering
\includegraphics[width=0.5\linewidth]{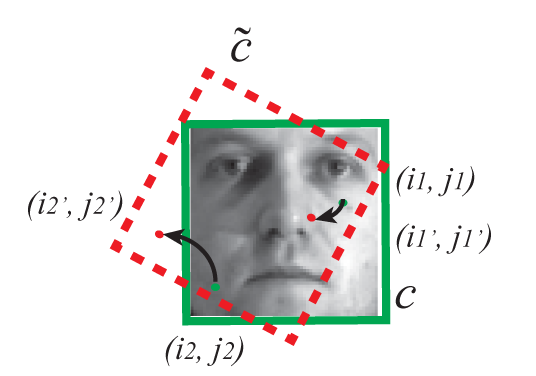}
\caption{Warping a cropped auxiliary image by $\tau_i^{-1}$ may result in copying some pixel values that are out of bound. The values of these out-of-bound pixels are not available in \eqref{eq:warp-a}. In this example, the pixel with the coordinates $(i_1', j_1')$ after transformation $\tau_i^{-1}$ remains within the original bounding box in green color, but $(i_2', j_2')$ is outside the original bounding box. Pixel coordinates such as $(i_2, j_2)$ should be removed from the support set $\Omega$.}
\label{fig:recognition-cropping}
\end{figure}

In addition, applying \eqref{eq:warp-a} that warps auxiliary images and gallery images to the query image sometimes can be undesirable in practice. Figure \ref{fig:recognition-cropping} illustrates the problem. In many cases, the auxiliary and gallery images are provided only within a cropped face region. Therefore, any pixel outside the original bounding box may not have a valid value. In some other cases, even when those pixels are available, still the original pixels within the training bounding box are typically well chosen to best represent the appearance of the face. As a result, using pixel values outside the bounding box may negatively affect the accuracy of the recognition.

In this paper, we propose a more efficient solution to address the problem. The key idea is to constrain the sparse representation-based classification on a subset of pixels whose pixel values remain valid after the alignment compensation \eqref{eq:warp-a}.

Without loss of generality, we assume each auxiliary image in $D$ is of dimension  $w\times h$, i.e., $d = wh$. In the SILT recognition step, given an estimated transformation from the alignment stage $\tau_i^{-1}$ for the gallery image $\aa_i$, we apply the transformation $\tau_i^{-1}$ on each pixel within the face image $(i,j)\in [1,w]\times[1,h]$. Define the support set for the transformation $\tau_i^{-1}$:
\begin{equation}
\Omega_i\doteq \{(i,j) |  \tau_i^{-1} (i,j) \in  [1,w]\times[1,h]  \}.
\end{equation}

Given all the collection of all the transformations $\tau_1, \tau_2, \ldots, \tau_L$, we define the total support set $\Omega$ as the intersection
\begin{equation}
\Omega = \bigcap_{i=1}^{L} \Omega_i,
\end{equation}
that is, each element in $\Omega$ corresponds to a valid pixel in the auxiliary images and $C$ after the transformations $\tau_i^{-1}$ are applied for all $i=1, \ldots, L$. The projection of an image in vector form $\bb$ onto a support set $\Omega$ is denoted as $\mathcal{P}_\Omega (\bb) \in \Re^{|\Omega|}$. 

\begin{figure}[ht!]
\includegraphics[width=\linewidth]{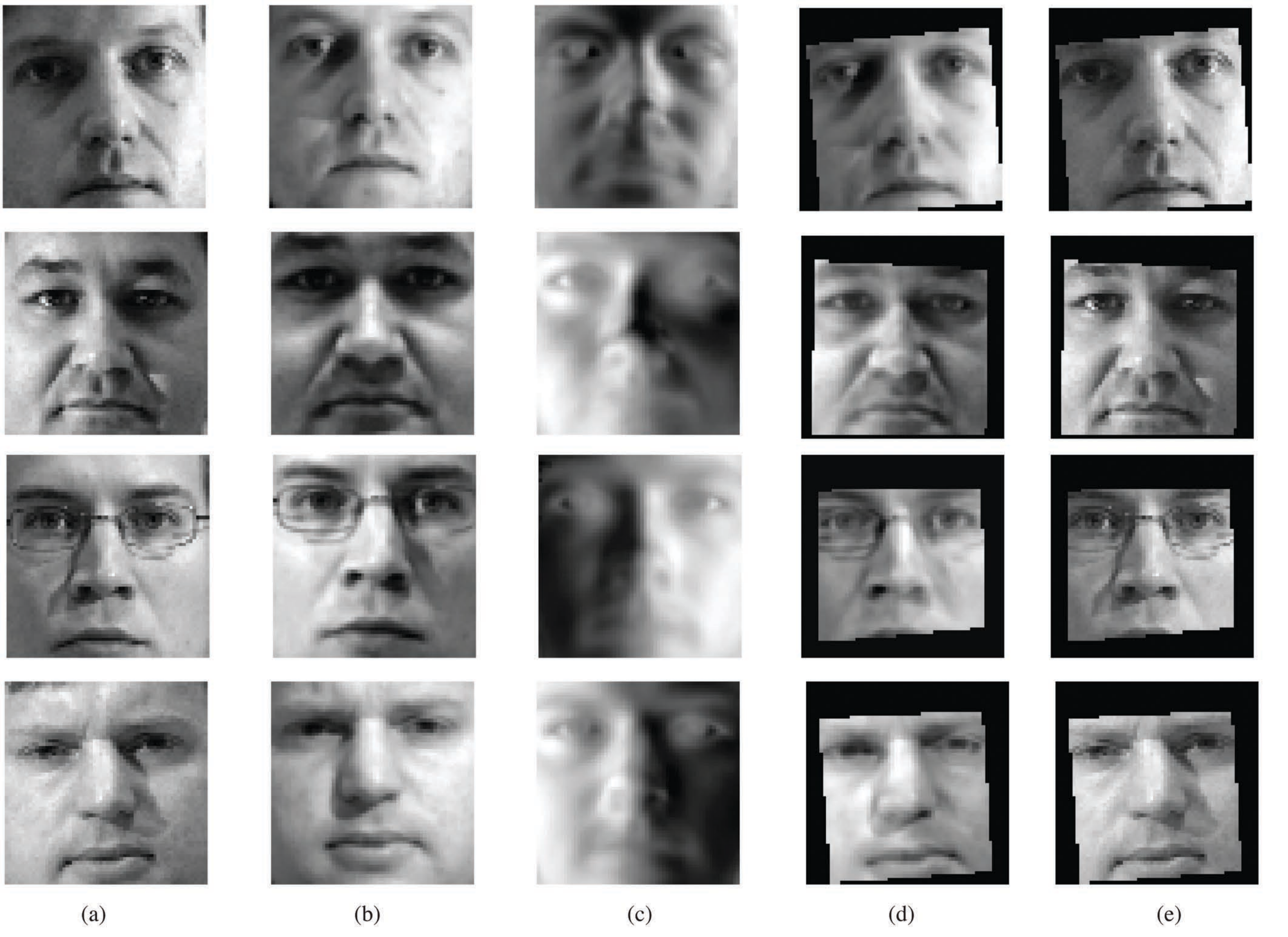}
\caption{Examples of warping a gallery image $\tilde{\aa}$ and applying a mask $\Omega$ on both the query image $\bb$ and the warped gallery image $\tilde{\aa}$. {\bf (a)}. Query images $\bb$. {\bf (b)}. Gallery images $\aa$. {\bf (c)}. Illumination transfer information $C\hat{\yy}$. {\bf (d)}. Warped gallery images $\tilde{\aa}$ under a mask $\Omega$. {\bf (e)}. Applying the same masks $\Omega$ on $\bb$. }
\label{fig:SILT-effect}
\end{figure}

The effect of applying a mask defined by a support set $\Omega$ is illustrated in Figure \ref{fig:SILT-effect}. Initially, the input query images in the first column and the gallery images of the same subjects have very different illumination conditions and poses. In the third column, an illumination transfer pattern is estimated for each gallery image. For example, in the second subject example, the left side of $\bb$ is brighter than that of $\aa$. This is reflected by having a brighter illumination pattern in its $C\hat{\yy}$. Finally, the gallery images are further warped based on the estimated poses $\tau^{-1}$, and the masks of their support sets $\Omega$ are applied to both the warped gallery images $\mathcal{P}_\Omega (\tilde{\aa})$ and the query images $\mathcal{P}_\Omega (\bb))$. We can see that, compared to the input image pairs $(\aa,\bb)$, the processed image pairs in the SILT algorithm $(\mathcal{P}_\Omega (\tilde{\aa}),\mathcal{P}_\Omega (\bb))$ have closer illumination conditions and similar poses.

The remaining SILT algorithm involves solving a sparse representation $\xx$ in the presence of a possible sparse error $\ee$ constrained on the support set $\Omega$, namely,
\begin{equation}
\begin{array}{rcl}
(\xx^*, \ee^*) &=& \arg\min_{\xx, \yy, \ee} \|\xx\|_1 + \lambda\|\ee\|_1 \\
\subjto & & \mathcal{P}_\Omega (\bb)  =  \mathcal{P}_\Omega (\tilde{A})\xx + \ee,
\end{array}
\label{eq:SILT-recognition}
\end{equation}
where the operation $\mathcal{P}_\Omega (\tilde{A})$ applies pixel selection on each column of $\tilde{A}$ based on the support set $\Omega$.
Similar to the previous formulations, the parameter $\lambda$ is chosen empirically via cross validation.

\begin{figure}[ht!]
\centering
{\includegraphics[width=\linewidth]{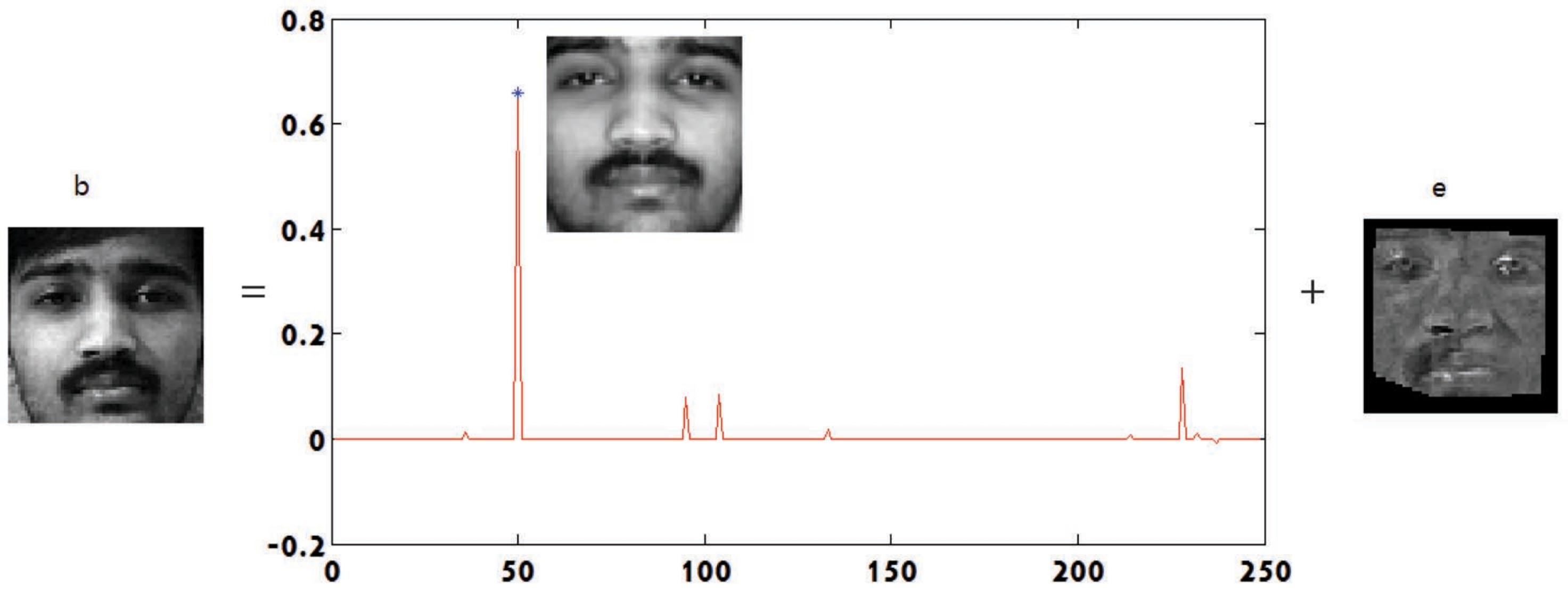}}
\caption{Illustration of SILT recognition. {\bf Left:} Query image $\bb$ with unknown pose and illumination. {\bf Right:} Sparse representation $\xx$ with the correct gallery image $\aa_i$ superimposed and sparse error $\ee$. The effect of pose alignment between $\bb$ and all the 250 gallery images is illustrated by the mask $\Omega$ shown in $\ee$.}
\label{fig:recognition}
\end{figure}

Using the sparse representation $\xx$ in \eqref{eq:SILT-recognition}, the final decision rule to classify $\bb$ can be simplified from the original SRC algorithm in \cite{WrightJ2009-PAMI} where the reconstruction residual was used. In SILT, since there is only one sample per each subject class in $A$, the class with the largest coefficient magnitude in $\xx$ is the estimated class of the query image $\bb$. We note that this simplified strategy does not compromise the generality of the SILT method, as one can still estimate the objective function of the reconstruction residual when each class contains one or more gallery images. Figure \ref{fig:recognition} shows an example of the SILT recognition and its estimated sparse representation.

Before we move on to examine the performance of the new recognition algorithm \eqref{eq:SILT-recognition}, one may question the efficacy of enforcing a sparse representation in the constraint \eqref{eq:SILT-recognition}. The question may arise because in the original SRC framework, the data matrix $A=\left[A_1,\cdots, A_L\right]$ is a collection of highly correlated image samples that span the $L$ illumination subspaces. Therefore, it makes sense to enforce a sparse representation as also validated by several followup studies \cite{WrightJ2010-IT,ElhamifarE2012-TSP,ZhangL2012}. However, in single-sample recognition, only one sample $\aa_i$ is provided per class. Therefore, one would think that the best recognition performance can only be achieved by the nearest-neighbor algorithm.

There are at least two arguments to justify the use of sparse representation in \eqref{eq:SILT-recognition}. One one hand, as discussed in \cite{WrightJ2009-PAMI}, in the case that $\ee$ and $C\yy$ represent a small error and the nearest-neighbor solution corresponds to a one-sparse binary vector $\xx_0=[\cdots,0,1,0 \cdots]^T$ in the formulation \eqref{eq:SILT-recognition}, then solving \eqref{eq:SILT-recognition} via $\ell_1$-minimization can also recover the sparsest solution, namely, $\xx^*\approx \xx_0$. On the other hand, in the case that $C\yy$ represents a large illumination change and $\ee$ represents additional gross image corruption, as long as the elements of $A$ in \eqref{eq:SILT-recognition} remain tightly correlated in the image space, the $\ell_1$-minimization algorithm can compensate the dense error in the query image $\bb$ \cite{WrightJ2010-IT}. This is a unique advantage over nearest-neighbor type algorithms.

\section{Experiment}
\label{sec:experiment}
In this section, we present a comprehensive experiment to demonstrate the performance of our illumination learning, face alignment, and recognition algorithms.

The illumination dictionary is constructed from Extended YaleB database \cite{LeeK2005-PAMI}. The Extended YaleB contains 21888 face image of 38 subjects under 9 poses and 64 illumination conditions. For every subject in a particular pose, an image with ambient (background) illumination was also captured. In this paper, only the frontal images of the 38 subjects are used as the auxiliary images.

For the gallery and query subjects, we choose images from a much larger CMU Multi-PIE database \cite{GrossR2006}. Except for Section \ref{sec:exp-robustness}, 166 shared subject classes from Session 1 and Session 2 are selected for testing. In Session 1, we randomly select one frontal image per class with arbitrary illumination as the gallery image. Then we randomly select two different frontal images from Session 1 or Session 2 for testing. The outer eye corners of both training and query images are manually marked as the ground truth for registration. All the training face images are manually cropped into $60 \times 60$ pixels based on the locations of eyes out-corner points, and the distance between the two outer eye corners is normalized to be 50 pixels for each person. We again emphasize that our experimental setting is more practical than those used in some other publications, as we allow the training images to have arbitrary illumination and not necessarily just the ambient illumination.

We compare our algorithms with several state-of-the-art face alignment and recognition algorithms under the SRC framework. To conduct a fair comparison, it is important to separate those algorithms that were originally proposed to handle only the recognition problem versus those that can handle both face alignment and recognition. The original SRC algorithm \cite{WrightJ2009-PAMI}, the Extended SRC (ESRC) \cite{DengW2012-PAMI}, and SVDL \cite{YangM2013-ICCV} belong to the first case, while Deformable SRC (DSRC) \cite{WagnerA2012-PAMI}, misalignment robust representation (MRR) \cite{YangM2012-ECCV}, and SILT proposed in this paper belong to the second case.

Finally, as the SILT algorithm relies on an auxiliary illumination dictionary $C$, another variability we need to investigate further is how the choice of $C$ may affect the performance of SILT. Our investigation on this issue will be divided in three steps. First, in Section \ref{sec: DL_simulations}, we validate in an ideal, noise-free simulation that the proposed dictionary learning algorithm can successfully recover the subject identity matrix $V$ and the illumination dictionary $C$ in \eqref{eq: modeling_ori}. We further utilize Extended YaleB database to construct an illumination dictionary from the real face images. Second, in Section \ref{sec:exp-robustness}, we will compare the recognition rates of SILT using different illumination dictionaries. The experiment further shows the SILT framework significantly outperforms DSRC and MRR in single-sample face recognition with misalignment and pixel corruption. Finally, in Section \ref{sec:exp-dictionary}, we again use Extended YaleB database to illustrate how the variation in the atom size and the training subjects of the auxiliary data affects the performance of the SILT algorithm.

\subsection{Learning Illumination Dictionaries}\label{sec: DL_simulations}
In this experiment, we validate the performance of the illumination dictionary learning algorithm in Algorithm \ref{algorithm_SOP}. First, we use noise-free synthetic data to evaluate the success rate for the algorithm to recover a subject-identity matrix $V$ and a sparsely-used dictionary $C$ as in \eqref{eq: L+BS}. Specifically, the elements in the $V\in\Re^{d\times p}$ and $C\in\Re^{d\times k}$ matrices are generated from independent and identically distributed (i.i.d.) Gaussian distributions. The columns of the sparse coefficient matrix $S\in \Re^{k\times np}$ are assumed to be $t$-sparse, where each column has exactly $t$ non-zero coefficients, where $n\doteq k\log_ek$ is the number of samples from each class and varies with the atom size $k$. These synthesized ground-truth matrices then generate the data matrices $D_1, D_2, \cdots, D_p \in \Re^{d\times n}$.

In the experiment, we set $d=100$, $p=5$, and let $k$ vary between 10 and 50 and $t$ between 1 and 10. In addition, to resolve the potential ambiguity in the permutation of the estimated dictionary atoms, we adopt the following relative error metric to a performance index:
\begin{equation}\label{SSE_a}
\phi(Z^*,Z)=\min_{\Pi,\Lambda} \|Z^*\Pi\Lambda-Z\|_F/\|Z\|_F
\end{equation}
where $\Pi$ is a permutation matrix, and $\Lambda$ is a diagonal scaling matrix.

\begin{figure}
\centering
\includegraphics[width=\linewidth]{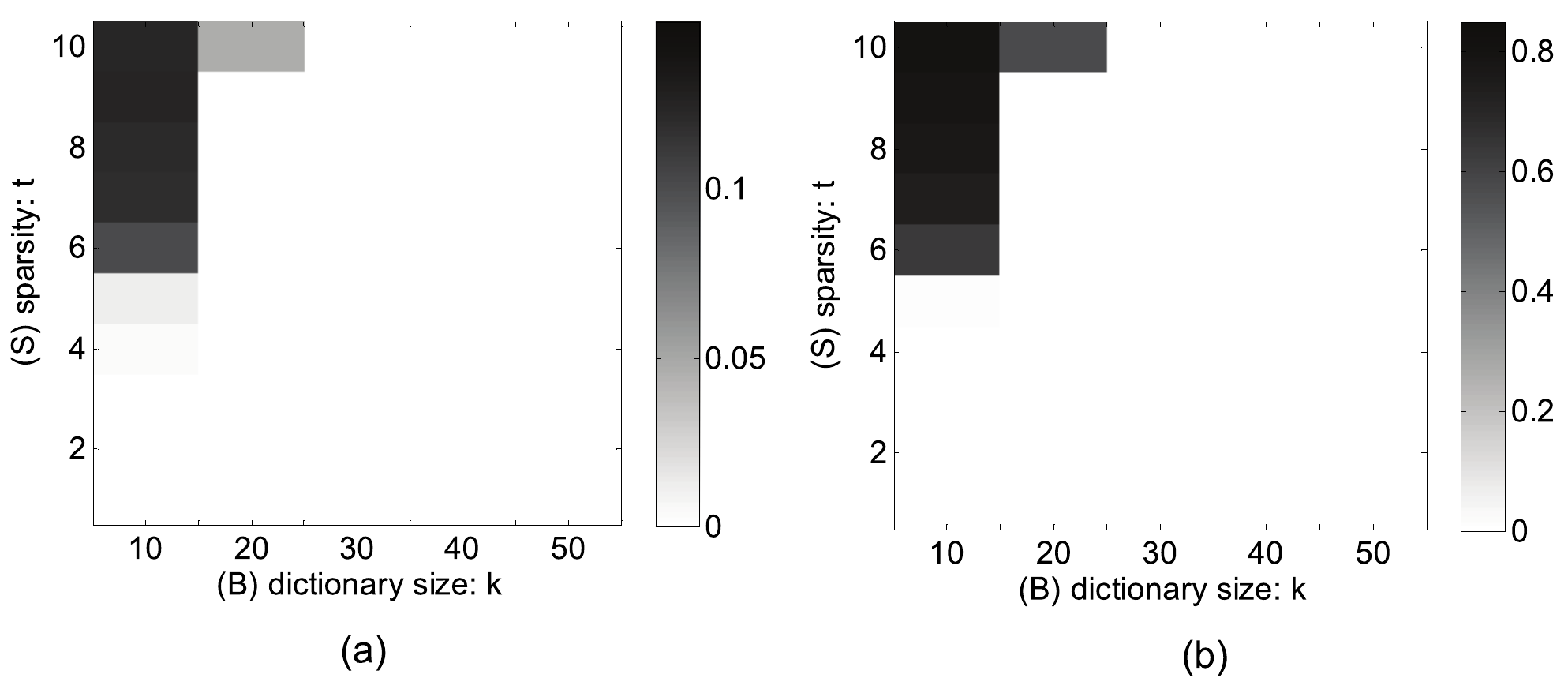}
\caption{Mean relative errors over 5 trials, with varying support $t$ and basis size $k$ for (a) $V$ and (b) $C$ estimated by Algorithm  \ref{algorithm_SOP}.}
\label{fig:mse}
\end{figure}

Figure \ref{fig:mse} shows the simulation result. The average relative error \eqref{SSE_a} for both $V$ and $C$ is reported in grayscale, where the white blocks indicate zero error, and the darker blocks indicate larger relative error. We can clearly see that when the dictionary size $k$ is sufficiently large and when the sparsity $t$ sufficiently small, Algorithm \ref{algorithm_SOP} perfectly recovers the two matrices. The algorithm only fails when $k=10, t<3$ and $k=20, t=10$. Furthermore, the phase transition from failed recovery settings to perfect recovery settings is quite sharp.

Next, we apply Algorithm \ref{algorithm_SOP} to learn the illumination dictionary $C$ from Extended YaleB database. For the experimental purpose in the subsequent sections, we construct two dictionaries with very different settings:
\begin{enumerate}
\item \emph{Ad-Hoc Dictionary}: We choose the very first subject in Extended YaleB database with 65 aligned frontal images (1 ambient + 64 illuminations). The dictionary $C$ is directly constructed by subtracting the ambient image from the other 64 illumination images, and no additional learning algorithm is involved. This dictionary is identical to the one used in our previous work \cite{Zhuang2013-CVPR}.
\item \emph{Yale Dictionary}: We employ all the 38 subjects in Extended YaleB database to learn an illumination dictionary using Algorithm \ref{algorithm_SOP}.
\end{enumerate}

\begin{figure}[ht!]
\centering
\includegraphics[width=0.98\textwidth]{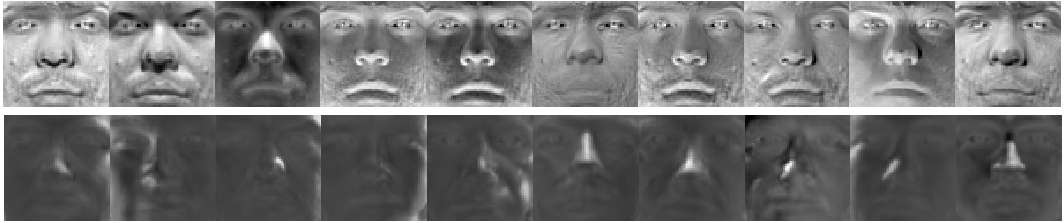}
\caption{Illustration of the first ten atoms of the illumination dictionary $C$. {\bf Top:} Ad-Hoc Dictionary constructed from the first subject of Extended YaleB database. {\bf Bottom:} Yale Dictionary learned from all the 38 subjects.}
\label{fig:dict}
\end{figure}

Some atoms from the above two dictionaries are shown as Figure~\ref{fig:dict}. The atom size of Yale Dictionary in this illustration is fixed at $80$. In Section~\ref{sec:exp-robustness} and \ref{sec:exp-dictionary}, we will compare the performance of different dictionaries.

\subsection{Simulation on 2D Alignment}
\label{sec:exp-alignment}
In this experiment, we demonstrate the performance of the SILT alignment algorithm \eqref{eq:SILT-alignment}. The performance is measured using simulated 2D deformation on the face image, including translation, rotation and scaling. Without loss of generality, we will only use Yale Dictionary as our illumination dictionary. The added deformation is introduced to the query images based on the ground truth coordinates of eye corners. The translation ranges from [-12, 12] pixels with a step size of 2 pixels.

Similar to \cite{WagnerA2012-PAMI}, we use the estimated alignment error $\|\ee\|_1$ as an indicator of success. More specifically, let $\ee_0$ be the alignment error obtained by aligning a query image from the manually labeled position to the training images. We consider the alignment successful if
$|\|\ee\|_1-\|\ee_0\|_1 | \leq 0.01\|\ee_0\|_1$.

We compare our method with DSRC and MRR. As DSRC and MRR would require to have multiple reference images per class,
to provide a fair comparison, we evaluate both algorithms under two settings: Firstly, seven reference images are provided per class to DSRC.\footnote{The training are illuminations \{0,1,7,13,14,16,18\} in Multi-PIE Session 1.} We denote this case as DSRC-7. Secondly, one randomly chosen image per class as the same setting as in the SILT algorithm. We denote this case as DSRC-1 and MRR-1, respectively.

\begin{figure*}[ht!]
\centering
\includegraphics[width=0.48\textwidth]{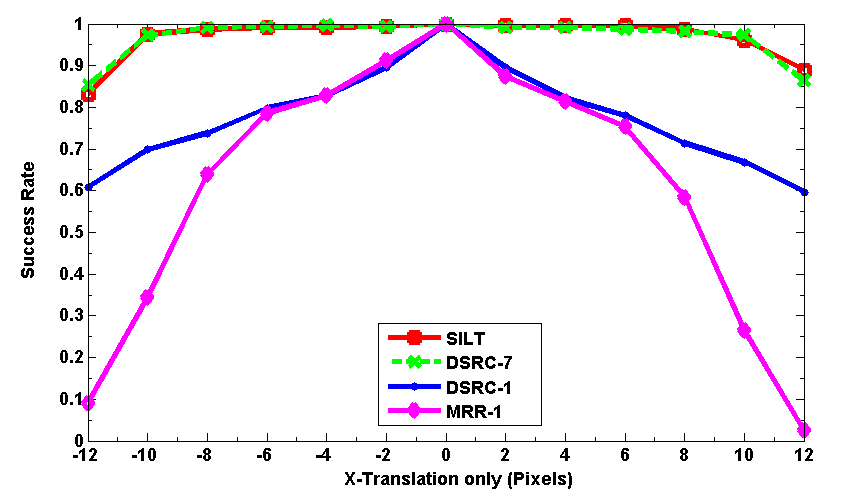}\quad
\includegraphics[width=0.48\textwidth]{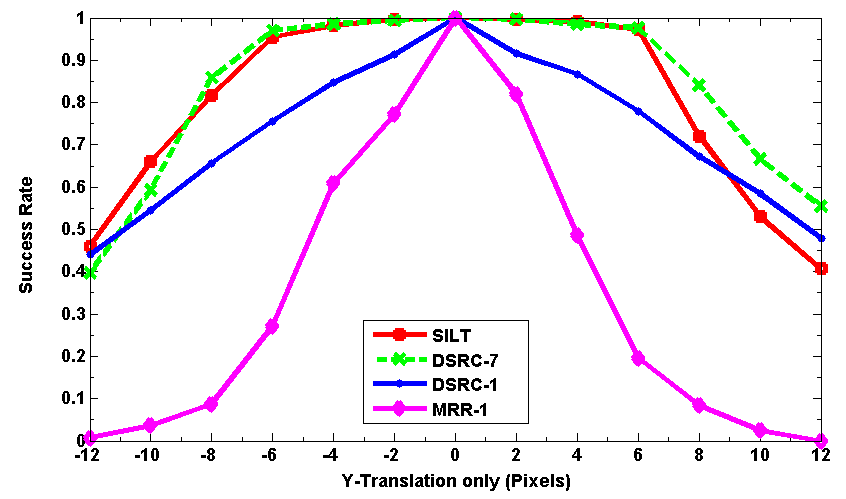}\\
\includegraphics[width=0.48\textwidth]{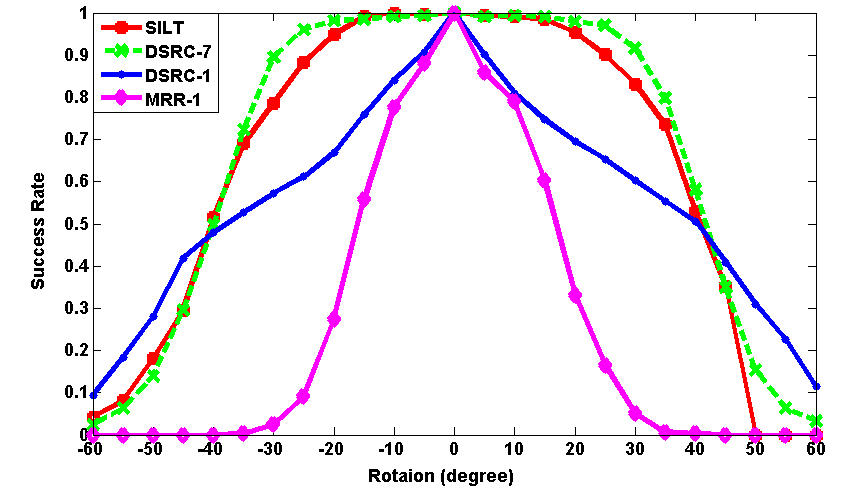}\quad
\includegraphics[width=0.48\textwidth]{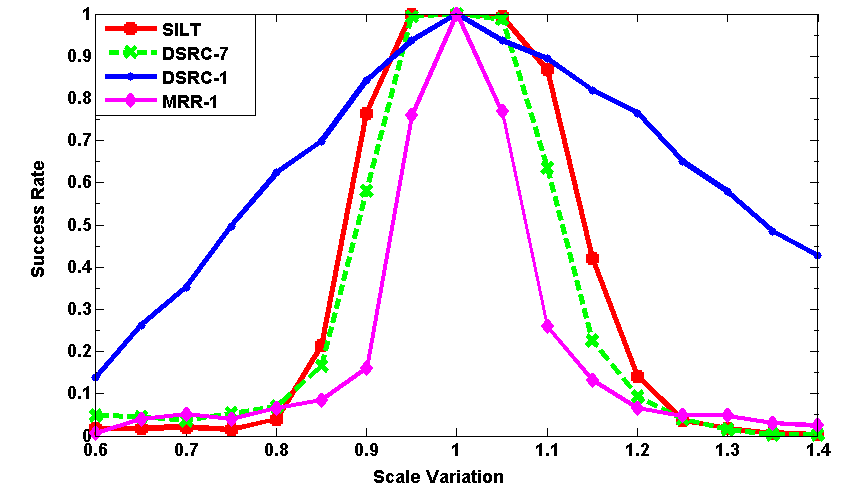}
\caption{Success rate of face alignment under four types of 2D deformation: $x$-translation, $y$-translation, rotation, and scaling. The amount of translation is expressed in pixels, and the in-plane rotation is expressed in degrees.
}
\label{fig:2d-deformation}
\end{figure*}

We draw the following observations from the alignment results shown in Figure~\ref{fig:2d-deformation}:
\begin{enumerate}
\item SILT works well under a broad range of 2D deformation, particularly when the translation in $x$ or $y$ direction is less than 20\% of the eye distance (10 pixels)
and when the in-plane rotation is less than 30 degrees.
\item Clearly, SILT outperforms both DSRC-1 and MRR-1 when the same setting is used, namely, one sample per class. The obvious reason is that DSRC and MRR were not designed to handle the single-sample alignment scenario.
\item The accuracy of SILT and DSRC-7 is generally comparable across the board in all the simulations. However, since DSRC-7 has access to seven gallery images of different illumination conditions, the result shows the power of using the new illumination dictionary in \eqref{eq:SILT-alignment}, where SILT only works with a single gallery image.
\end{enumerate}

\subsection{Single-Sample Recognition}
\label{sec:exp-recognition}

In this subsection, we evaluate the SILT recognition algorithm based on single reference images of the 166 subject classes shared in Multi-PIE Sessions 1 and 2. We compare its performance with SRC \cite{WrightJ2009-PAMI}, ESRC \cite{DengW2012-PAMI}, DSRC \cite{WagnerA2012-PAMI}, MRR \cite{YangM2012-ECCV}, and SVDL \cite{YangM2013-ICCV}. The illumination dictionary used in these experiments is Yale Dictionary.

First, we note that the new SILT framework and the existing sparse representation algorithms are \emph{not} mutually exclusive. In particular, the illumination transfer \eqref{eq:warp-a} can be easily adopted by the other algorithms to improve the illumination condition of the training images, especially in the single-sample setting. In the first experiment, we demonstrate the improvement of SRC and ESRC with the illumination transfer. Since both algorithms do not address the alignment problem, manual labels of the face location are assumed to be the aligned face location. The comparison is presented in Table \ref{table:manual_recog_rate}.

\begin{table}[ht!]
\centering
\caption{Single-sample recognition accuracy via manual alignment. The atom size is fixed to 80.}
\begin{tabular}{|l|cc|}
\hline
Method           &  Session 1 (\%) & Session 2 (\%)\\
\hline
\hline
$\mbox{SRC}_M$  &  88.0 & 53.6  \\
$\mbox{ESRC}_M$  &  89.6 & 56.6  \\
\hline
SILT + $\mbox{SRC}_M$ &  92.8 & 59.0  \\
SILT + $\mbox{ESRC}_M$ &  \textbf{93.2} & \textbf{59.3}  \\
\hline
$\mbox{SVDL}_M$  &  70.3 & 41.6 \\
\hline
\end{tabular}
\label{table:manual_recog_rate}
\end{table}

Since the gallery images are selected from Session 1, there is no surprise that the average recognition rate of Session 1 is significantly higher than that of Session 2. The comparison further shows that adding the illumination transfer information to the existing SRC and ESRC algorithms meaningfully improves their performance by 3\% -- 5\%.

In Table \ref{table:manual_recog_rate}, the performance of the SVDL algorithm is also shown.\footnote{The implementation of SVDL was provided by their authors at: {http://www4.comp.polyu.edu.hk/~cslzhang/code/SVDL.zip}.} Interestingly, in our setting of single-sample recognition, SVDL performs worse than SRC and ESRC. A possible explanation is that the SVDL algorithm expects all the gallery images to have the same uniform lighting condition, while in this paper, the illumination condition of the gallery images is randomly selected. Our experimental setting is more challenging but more similar to the single-sample face recognition problem in practice. Furthermore, one can consider combining the SILT framework and the illumination dictionary of SVDL. This variation will be considered in Section \ref{sec:exp-robustness}.

Second, we compare DSRC, MRR, and SILT in the full pipeline of alignment plus recognition shown in Table \ref{table:recog_rate}. The initial positions of the face images are automatically detected by Viola-Jones detector.
\begin{table}[ht!]
\centering
\caption{Single-sample alignment + recognition accuracy.}
\begin{tabular}{|l|cc|}
\hline
Method           &  Session 1 (\%) & Session 2 (\%)\\
\hline
\hline
DSRC             & 36.1 & 35.7  \\
MRR              &  46.2 & 34.6  \\
\hline
SILT       & \textbf{76.7} & \textbf{61.6} \\
\hline
\end{tabular}
\label{table:recog_rate}
\end{table}

Compared with the past reported results of DSRC and MRR, their recognition accuracy decreases significantly when only one training image is available per class. It demonstrates that these algorithm were not designed to perform well in the single-sample regime. In both Session 1 and Session 2, SILT outperforms both algorithms by more the 30\%. It is more interesting to compare the recognition rates of different algorithms on Session 2 in Table \ref{table:manual_recog_rate} and Table \ref{table:recog_rate}. SILT that relies on an auxiliary illumination dictionary to automatically alignment the query images achieves 61.6\%, which is even higher than the ESRC rate of 59.3\% with manual alignment.

\subsection{Robustness under Random Corruption}
\label{sec:exp-robustness}
In this subsection, we further compare the robustness of the SILT recognition algorithm to random pixel corruption. We compare the overall recognition rate of SILT with DSRC, and MRR, the two most relevant algorithms. For the SILT algorithm, in addition to using the two previous illumination dictionaries, namely, Ad-Hoc and Yale, we also demonstrate the performance using the SVDL dictionary \cite{YangM2013-ICCV}.

To benchmark the recognition under different corruption percentage, it is important that the query images and the gallery images have close facial appearance, otherwise different facial features would also contribute to facial corruption or disguise, such as glasses, beard, or different hair styles. To limit this variability, in this experiment, we select both query and gallery images from Multi-PIE Session 1, although the images should never overlap. We use all the subjects in Session 1. For each subject, we randomly select one frontal image with arbitrary illumination for testing. Various levels of image corruption from 10\% to 40\% are randomly generated in the face region. Similar to the previous experiments, the face regions are detected by Viola-Jones detector. The performance is shown in Table \ref{table:robust_rate}.

\begin{table}[htbp]
\centering
\caption{Recognition rates (\%) under various percentage of random pixel corruption. The atom size is fixed to 80.}
\begin{tabular}{|l|cccc|}
\hline
Corruption   &  10\%  & 20\%  &  30\%   &    40\% \\
\hline
\hline
DSRC & 32.9 & 31.7 & 28.9 & 24.1 \\
MRR & 24.9 & 14.5 & 11.7 & 9.2 \\
\hline
$\mbox{SILT}(\mbox{Ad-Hoc})$ & 66.2 & 59.8 & 49.6 & 44.7 \\
$\mbox{SILT}(\mbox{Yale})$ & {\bf 73.3} & {\bf 68.7} & {\bf 67.3} & {\bf 49.0} \\
$\mbox{SILT}(\mbox{SVDL})$ & 60.0 & 56.1 & 52.3 & 41.1 \\
\hline
\end{tabular}
\label{table:robust_rate}
\end{table}

The comparison is more illustrative than Table \ref{table:recog_rate}. First of all, all three SILT implementations based on very different illumination dictionaries significantly outperform DSRC and MMR. For instance, with 40\% pixel corruption, SILT still maintains 49\% accuracy; with 10\% corruption, SILT outperforms DSRC and MRR by more than 40\%. 

Second, we note that in the presence of pixel corruption, the illumination dictionary learned by SVDL does not perform as well as Ad-Hoc and Yale dictionaries. It shows that our proposed dictionary learning method is more suited for estimating auxiliary illumination dictionaries in the SILT framework.

\subsection{Influence of Atom Size and Subject Number}
\label{sec:exp-dictionary}

In this section, we discuss how the efficacy of an SILT dictionary may be affected by the choice of the atom size and the subject number. More specifically, We learn illumination dictionaries using Algorithm \ref{algorithm_SOP} from Extended YaleB database with varying number of the auxiliary subjects and atom size of the dictionary. Then, we measure the accuracy of face recognition under the frameworks of ``SILT+ESRC$_M$" and ``SILT+SRC$_M$" with manual alignment. The settings is the same as Section~\ref{sec:exp-recognition}, namely, gallery and query images are chosen from Session 1 of Multi-PIE database. The results are shown in Table \ref{table:atom_size_ESRC} and Table \ref{table:atom_size_SRC}.
\begin{table}[htbp]
\centering
\caption{Recognition rates (\%) under the SILT+ESRC$_M$ implementation with manual alignment.}
\begin{tabular}{|l|ccccc|}
\hline
atom size   &  40  & 60  &  80   & 120 &  200  \\
\hline
\hline
subject \# = 1  & 89.6 & 89.2 & 89.2 & 89.2 & - \\
subject \# = 10 & 90.0 & 92.8 & 92.8 & 94.0 & 94.8 \\
subject \# = 38 & 90.8 & 93.2 & 93.2 & 95.2 & {\bf 96.8} \\
\hline
\end{tabular}
\label{table:atom_size_ESRC}
\end{table}
\begin{table}[htbp]
\centering
\caption{Recognition rates (\%) under the SILT+SRC$_M$ implementation with manual alignment.}
\begin{tabular}{|l|ccccc|}
\hline
atom size   &  40  & 60  &  80   & 120 &  200  \\
\hline
\hline
subject \# = 1  & 86.8 & 88.0 & 87.2 & 87.2 & - \\
subject \# = 10 & 87.2 & 91.2 & 92.4 & 90.8 & 92.8 \\
subject \# = 38 & 91.2 & 93.2 & 92.8 & 94.8 & {\bf 95.6} \\
\hline
\end{tabular}
\label{table:atom_size_SRC}
\end{table}

First, we notice that there is no data point taken at 200 atom size when the subject number is one. This is due to the fact that each subject in Extended YaleB database only provides 65 frontal images. When one tries to solve for more atoms in the corresponding illumination dictionary in \eqref{eq: modeling_ori}, the problem becomes ill-conditions. This issue can be first observed by examining the recognition rates for one subject and atom sizes greater than 60, namely, 80 and 120. In these two settings, the recognition rates are either identical or slightly worse than those at atom size 60 in both Table \ref{table:atom_size_ESRC} and Table \ref{table:atom_size_SRC}. At atom size 200, through visual inspection, we discover that the illumination patterns in the estimated $C$ matrices are close to random noise, and do not contain useful illumination information for the SILT algorithm. Therefore, their performance is ignored.

Second, when the subject number is higher than one, increasing the atom size of the illumination dictionary clearly improves the recognition rate. For example, using all the 38 subjects and the SILT+ESRC$_M$ algorithm, the recognition rate using a 40-atom illumination dictionary is 90.8\%. The rate is raised to 96.8\% when the atom size increases to 200. It is worth emphasizing that this recognition rate represents one of the best accuracy on Multi-PIE database when only single gallery images of random illumination are available, to the best of our knowledge.

Finally, it comes as no surprise that if we fix the size of the illumination dictionary in each column of Table \ref{table:atom_size_ESRC} and Table \ref{table:atom_size_SRC}, including more subjects in the illumination database also improves the recognition. This phenomenon can be explained by considering the well-known Lambertian model of the human face. It states that the image appearance of a face is determined not only by the illumination of the environment, but also by the shape of the face and its surface albedo pertaining to individual subjects. Therefore, having more subjects would help to generalize the distribution of the illumination patterns under different face shape and albedo. Then, the use of sparse representation in the alignment and recognition algorithms can effectively select a sparse subset of these illumination patterns that are most similar to the illumination, shape, and albedo condition of the query image.

\section{Conclusion and Discussion}
\label{sec:conclusion}

In this paper, we have presented a novel face recognition algorithm specifically designed for single-sample alignment and recognition. To compensate for the missing illumination information traditionally provided by multiple gallery images, we have proposed a novel dictionary learning algorithm to estimate an illumination dictionary from auxiliary training images. We have further proposed an illumination transfer technique to transfer the estimate illumination compensation and pose information from the face alignment stage to the recognition stage. The overall algorithm is called \emph{sparse illumination learning and transfer} (SILT). The extensive experiment has validated that not only the standalone SILT algorithm outperforms the state of the art in single-sample face recognition by a significant margin, the illumination learning and transfer technique is also complementary to many existing algorithms as a pre-processing step to improve the image condition due to misalignment and pixel corruption.

Although we have provided some exciting results that represent a meaningful step forward towards a real-world face recognition system in this paper, one of the open problems remains to be how to improve illumination transfer in complex real-world conditions and with minimal training data. Although the current way of constructing the illumination dictionary is efficient, the method is not able to separate the effect of surface albedo, shape, and illumination completely from face images. Therefore, we believe a more sophisticated illumination transfer algorithm could lead to better overall performance.


\section*{Appendix}
We proof Theorem \ref{thm} in this appendix. First, eliminating the variable $E$ of problem \eqref{eq: successive1} with
\begin{equation}\label{eq: E}
E= {D} - \overline{{V}}\otimes{\mathbf
1}^T - \overline{{C}}{H},
\end{equation}
Problem \eqref{eq: successive1} can then be equivalently written as
\begin{equation}\label{eq: modeling_322}
\begin{split} \min_{\overline{{V}},\overline{{C}}, {H}} & ~~\|{D} - \overline{{V}}\otimes{\mathbf 1}^T -
\overline{{C}}{H}\|_F^2,~{\rm s.t.~} \overline{{C}}^T\overline{{C}} =
{I}.
\end{split}
\end{equation}
As a basic result in least squares \cite{Horn1985}, the optimal ${H}$ can be
written as
\begin{equation}\label{eq: X}
{H}^* = \overline{{C}}^T({D} - \overline{{V}}\otimes{\mathbf 1}^T),
\end{equation} for any $\overline{{V}}\in \Re^{m\times p}$ and any $\overline{{C}}\in \Re^{m\times k}$ such that $\overline{{C}}^T\overline{{C}} = {I}$. Substituting ${H}^*$ into \eqref{eq: modeling_322}
yields
\begin{equation}\label{eq: modeling_32211}
\begin{split} \min_{\overline{{V}},\overline{{C}}} & ~~\|{P}_{\overline{{C}}}^\perp ({D} - \overline{{V}}\otimes{\mathbf 1}^T)\|_F^2,~{\rm s.t.~} \overline{{C}}^T\overline{{C}} = {I},
\end{split}
\end{equation}where ${P}_{\overline{{C}}}^\perp = {I} - \overline{{C}}\overline{{C}}^T$ denotes the orthogonal complement projector of $\overline{{C}}$.
It is also easy to show from \eqref{eq: modeling_32211} that a
solution of $\overline{{V}}$ is
\begin{equation}\label{eq: A_bar_est}
[\overline{{V}}^*]_i = \frac{1}{n} {D}_i{\mathbf 1},~i=1,...,p.
\end{equation}Note that the solution $[\overline{{V}}^*]_i$ presents
the mean vector of the data matrix ${D}_i$ corresponding to subject $i$. Furthermore, by letting
${U} = {D} - \overline{{V}}^*\otimes{\mathbf 1}^T$, problem \eqref{eq: modeling_32211} becomes
$\min_{\overline{{C}}^T\overline{{C}} = {I}} {\rm
trace}({U}^T{P}_{\overline{{C}}}^\perp{U})$, and it is equivalent to
\begin{equation}
\overline{{C}}^* = {\rm arg}\max_{\overline{{C}}^T\overline{{C}} = {I}} {\rm trace}(\overline{{C}}^T{U}{U}^T\overline{{C}}).
\end{equation}By \cite{Horn1985}, an optimal solution $\overline{{C}}^*$ is known to
be the $k$ principal eigenvector matrix of ${U}{U}^T$; i.e.,
\begin{equation}\label{eq: B_bar_est}
\overline{{C}}^* = [~ \boldsymbol{q}_1( {U}{U}^T ),
\boldsymbol{q}_2( {U}{U}^T ), \hdots, \boldsymbol{q}_{k}(
{U}{U}^T ) ~].
\end{equation}
Hence, the problem solution \eqref{eq: PCA} simply follows from \eqref{eq: E}, \eqref{eq: X} \eqref{eq: A_bar_est}, and \eqref{eq: B_bar_est}. \hfill
$\blacksquare$

\end{document}